\documentclass{ieeeaccess}
\usepackage{cite}
\usepackage{amsmath,amssymb,amsfonts}
\usepackage{algorithmic}
\usepackage{graphicx}
\usepackage{textcomp}
\usepackage{pifont}
\usepackage{multirow}
\usepackage{capt-of}
\usepackage{hyperref}

\usepackage{bm}
\makeatletter
\AtBeginDocument{\DeclareMathVersion{bold}
\SetSymbolFont{operators}{bold}{T1}{times}{b}{n}
\SetSymbolFont{NewLetters}{bold}{T1}{times}{b}{it}
\SetMathAlphabet{\mathrm}{bold}{T1}{times}{b}{n}
\SetMathAlphabet{\mathit}{bold}{T1}{times}{b}{it}
\SetMathAlphabet{\mathbf}{bold}{T1}{times}{b}{n}
\SetMathAlphabet{\mathtt}{bold}{OT1}{pcr}{b}{n}
\SetSymbolFont{symbols}{bold}{OMS}{cmsy}{b}{n}
\renewcommand\boldmath{\@nomath\boldmath\mathversion{bold}}}

\newcommand{\projname}[1]{{L2CU}}
\newcommand{\modelname}[1]{{AI cooperative model}}
\makeatother

\def\BibTeX{{\rm B\kern-.05em{\sc i\kern-.025em b}\kern-.08em
    T\kern-.1667em\lower.7ex\hbox{E}\kern-.125emX}}

\begin{document}
\history{Date of publication xxxx 00, 0000, date of current version xxxx 00, 0000.}
\doi{10.1109/ACCESS.2024.0429000}

\title{\projname{}: Learning To Complement Unseen Users}
\author{\uppercase{Dileepa Pitawela}\authorrefmark{1},
\uppercase{Gustavo Carneiro}\authorrefmark{2},
\uppercase{Hsiang-Ting Chen}\authorrefmark{1}
}

\address[1]{The School of Computer Science, University of Adelaide, Australia}
\address[2]{Centre for Vision, Speech and Signal Processing, University of Surrey, UK}

\markboth
{Pitawela \headeretal: \projname{}: Learning To Complement Unseen Users}
{Pitawela \headeretal: \projname{}: Learning To Complement Unseen Users}

\corresp{Corresponding author: Hsiang-Ting Chen (tim.chen@adelaide.edu.au)}

\begin{abstract}
Recent research highlights the potential of machine learning models to learn to complement (L2C) human strengths; however, generalizing this capability to unseen users remains a significant challenge.
Existing L2C methods oversimplify interaction between human and AI by relying on a single, global user model that neglects individual user variability, leading to suboptimal cooperative performance.
Addressing this, we introduce \projname{}, a novel L2C framework for human-AI cooperative classification with unseen users. Given sparse and noisy user annotations, \projname{} identifies representative annotator profiles capturing distinct labeling patterns.
By matching unseen users to these profiles, \projname{} leverages profile-specific models to complement the user and achieve superior joint accuracy.
We evaluate \projname{} on datasets (CIFAR-10N, CIFAR-10H, Fashion-MNIST-H, Chaoyang and AgNews), demonstrating its effectiveness as a model-agnostic solution for improving human-AI cooperative classification.
\end{abstract}


\titlepgskip=-21pt

\maketitle

\section{Introduction}
\label{sec:introduction}
\PARstart{H}{uman}-AI cooperation aims to combine the strengths of humans and AI to achieve superior performance compared to either acting alone.  Within this field, learning to defer (L2D) and learning to complement (L2C) represent distinct approaches.
L2D focuses on AI abstention when confidence is low, relying on human intervention for difficult cases ~\cite{predictResponsibly_madras,whoshould_mozannar23,fifar}.
L2C, however, aims for a more synergistic partnership, where both human and AI actively contribute their complimentary strengths, leading to greater overall performance ~\cite{complement_wilder,Bayesian_Steyvers22}.
Both L2D and L2C are vital strategies for effective human-AI cooperative tasks.
L2D prioritizes efficient use of human time by focusing human expertise on instances where AI is uncertain, whereas L2C priorities maximal joint accuracy through the human-AI cooperation. 

\begin{figure}[t]
    \centering
    \includegraphics[width=\columnwidth]{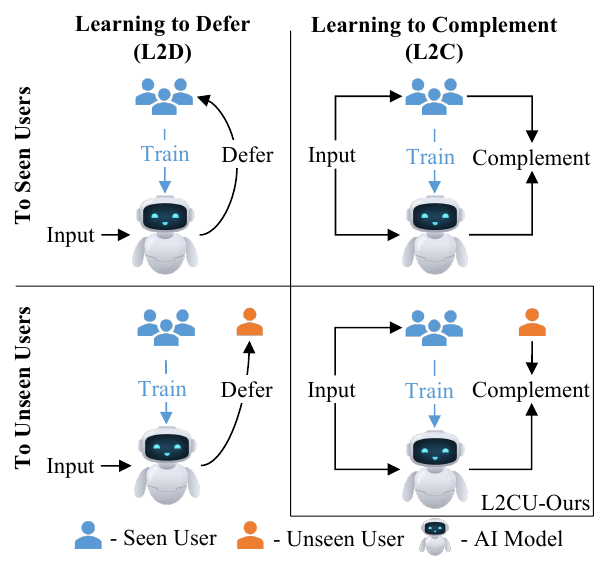}
    \caption{Paradigms of Human-AI Cooperation with seen users in blue and unseen users in orange.
    L2D defers decisions to humans, evolving to handle both seen and unseen users.
    L2C complements human strengths, and \projname{} (Ours) advances L2C to complement unseen users.
    }
    \label{fig:teaser}
\end{figure}

A key challenge in deploying human-AI cooperative systems is ensuring its effective generalization to diverse skills and behaviors of unseen users (i.e., drawn from the same distribution as users in the training set but not included in it).
Recent advances in L2D begin to address this challenge. 
For example, EA-L2D ~\cite{ea_l2d} employs a Bayesian approach for expert-agnostic modeling and generalizes to unseen experts. L2D-Pop ~\cite{l2d_2pop} utilizes meta-learning to adapt to new experts.
In contrast, the challenge of complementing unseen users in L2C remains largely under-explored (Figure~\ref{fig:teaser}).
The task is inherently more complex than L2D's binary deferral because L2C requires adapting a model to latent user characteristic labeling patterns based on sparse and potentially incorrect labels, given that such sparsity and noise are inevitable in real-world datasets.

\begin{figure*}[t]
    \centering
    \includegraphics[width=\textwidth]{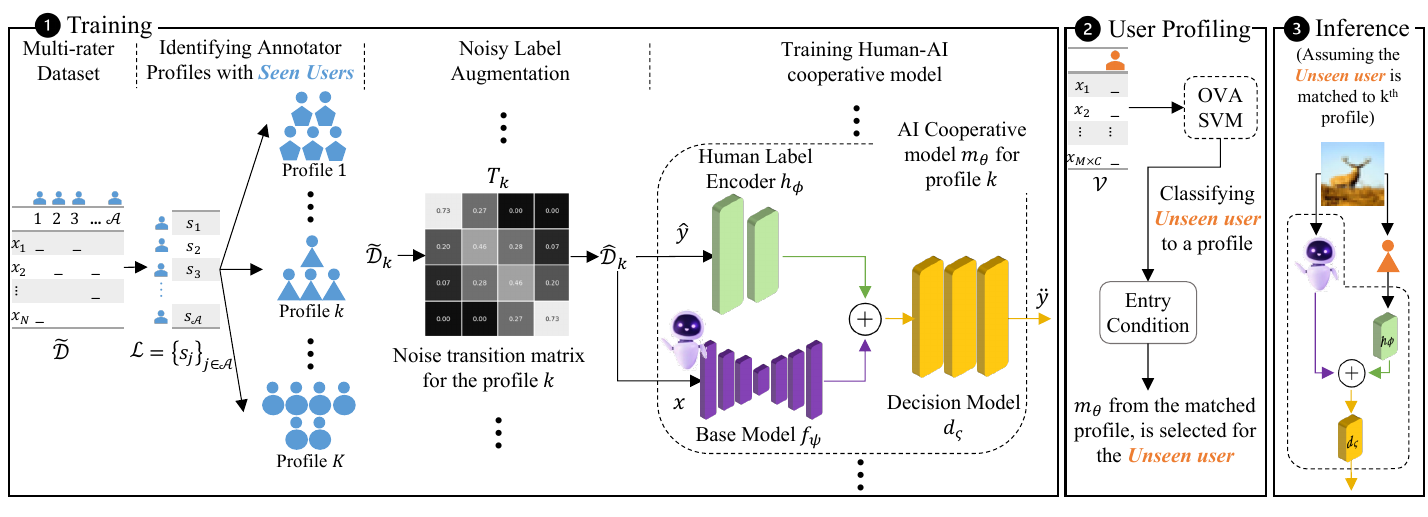}
    \caption{Three step \projname{} framework. 1) During training, from a sparse multi-rater dataset, unique annotator profiles are identified $(1,...,K)$. Then, for each annotator profile, noisy label augmentation is performed and a \modelname{} is trained. 2) During user profiling, a test user annotates a validation set and based on validation labels, a profile is matched by OVA SVM, entry condition is evaluated, and respective \modelname{} is selected. 3) At inference, the test user is paired with the corresponding model from the selected profile for cooperative classification. 
    }
\label{fig:framework}
\end{figure*}

To address this research gap, this paper introduces \projname{}, a novel L2C framework for human-AI cooperative classification, designed to achieve optimal performance with unseen users (Figure~\ref{fig:framework}).
Given a sparse training dataset with noisy labels, where training users only label subset of samples, \projname{} extracts representative annotator profiles, each capturing a distinct noisy labeling pattern. 
These profiles are then used to train \modelname{} instances, each optimized for a specific profile. 
To enable cooperation at test time, a user profiling process matches a new user to a profile, and the corresponding \modelname{} is selected. 
This profiling mechanism enables \projname{} to generalize to unseen users who were not included in training, and adapt to their characteristic labeling patterns.

We thoroughly evaluate \projname{} on both simulated and real multi-rater settings across diverse modalities (image and text) and domains (everyday objects, news classification, and medical diagnosis), with CIFAR-10N, CIFAR-10H, Fashion-MNIST-H, Chaoyang and AgNews datasets. We also introduce a novel assessment metric, alteration rate, which quantifies the extent to which the model's predictions improve upon or deviate from the original human labels. Our results demonstrate that \projname{}, a model-agnostic L2C framework, generalizes effectively to unseen users, consistently outperforming both individual human annotators and leading L2D and L2C methods across a range of classification tasks.
Our key contributions include:
\begin{itemize}
\item \projname{}, a new learning-to-complement framework designed to complement users unseen during training.

\item \projname{} handles sparse, multi-user settings and proposes a label augmentation method to augment sparse training data while preserving characteristic labeling patterns.

\item \projname{} achieves leading performance in human-AI cooperative classification and introduces a novel alteration rate metric to offer insights into the model's impact on human labels.
\end{itemize}

\projname{}'s model-agnostic design, ability to train using sparse noisy labels (without accessing ground truth), and generalization to unseen users make it a significant
contribution to the field of human-AI cooperative classification
\footnote{Code is available at: \href{https://github.com/dpitawela/L2CU}{https://github.com/dpitawela/L2CU}}.

\section{Related Work}
The uncertainties of automation often demand human involvement, leading to new human-AI cooperation paradigms ~\cite{Strauch2018-IronyAutomation}.
Learning-to-Defer (L2D) methods let AI models handle confident cases while deferring uncertain ones to humans by optimizing a utility function that balances model accuracy, preference of a human decision, and deferral costs ~\cite{predictResponsibly_madras,whoshould_mozannar23,fifar}. 
For instance, \cite{raghu19_triage} used AI model ensemble to flag high-risk patients for human review.
\cite{okati21DiffTriage} refines classifiers and use post-hoc rejectors to identify when to defer.
\cite{consistentest_Mozannar2020, verma22OVA, whoshould_mozannar23} optimize surrogate loss functions for deferral.
\cite{prob_l2d} optimizes work distribution for deferral.
However, these assume clean labels or defer only to seen users.
In contrast, we propose a learning-to-complement approach that removes the need for clean labels and supports cooperation with unseen users.

Learning-to-Defer to Unseen Users (L2DU) aims to defer to test users different from the training users.
EA-L2D \cite{ea_l2d} proposes a Bayesian framework to model expert behavior without expert-specific training data. With this prior knowledge about experts, they generalize to unseen experts.
L2D-Pop \cite{l2d_2pop} uses a small context set to characterize an expert's decision patterns and employs meta-learning to refine deferral policies dynamically.
However, it assumes each training user labels all training samples—uncommon in multi-rater settings.
In contrast, being a L2C method, \projname{} addresses realistic conditions where each training user labels only a subset of the data.

Learning-to-Complement (L2C) is less explored than L2D.
L2C leverages the strengths of both humans and AI to improve decision-making.
\cite{complement_wilder} considers uncertainty from both sides to improve decision, \cite{Bayesian_Steyvers22} uses Bayesian modeling for human-AI complementary.
LECOMH \cite{l2cmh} estimates human-AI consensus, then trains a selection module to minimize error and collaboration cost.
LECODU \cite{l2cdmu} further decides when to collaborate or defer, and how many experts to involve.
However, unlike those methods that complement the same users appearing in training, \projname{} enables complementing unseen users.

While adapting to users is studied in recommender systems, the absence of user-item preference data and L2C's distinct goal of complementing user biases to improve joint accuracy—unlike predicting preferences—render such methods inapplicable, leaving user adaptation in L2C underexplored \cite{Kocielnik2019-Imperfect,Wang2021-Brilliant}.
Available L2C work model users globally using confusion matrices \cite{Kerrigan2021-calibrateconfmatrix} or behavior models \cite{Vodrahalli2022_uncalibModels}, but these overlook individual biases and require extensive human labels.
In contrast, \projname{} identifies distinct annotator profiles and adapts without requiring extra labels to train behavior models.
Appendix~\ref{supp:related_work_ex} extends the related work.

\section{Methodology}
\subsection{Problem Formulation}
\label{sec:notation}

We consider a classification problem with $C$ classes, where each data sample has multiple, potentially noisy labels provided by different annotators. 
Our training dataset, denoted as  $\tilde{\mathcal{D}}=\{ (\mathbf{x}_i,\{\tilde{\mathbf{y}}_{i,j}\}_{j \in \mathcal{A}}) \}_{i=1}^{N}$, consists of $N$ data samples.  Each sample $\mathbf{x}_i \in \mathcal{X}$ has a set of noisy labels $\{\tilde{\mathbf{y}}_{i,j}\}_{j \in \mathcal{A}}$, where $\tilde{\mathbf{y}}_{i,j} \in \{0, 1, ..., C-1\}$ represents the label provided by annotator $j$ for sample $i$.  The set $\mathcal{A}$ represents the set of all annotators. Crucially, each annotator labels only a \textit{subset} of the samples. 
We assume each sample $\mathbf{x}_i$ has a latent, clean label $\mathbf{y}_i$, and that annotator noise is class-dependent \cite{song2022learning}. 
When clean labels are unavailable, we employ Crowdlab \cite{crowdlab} to estimate consensus labels, $\bar{\mathbf{y}}_i$, for each sample, resulting in a consensus-labeled dataset $\bar{\mathcal{D}} = \{ (\mathbf{x}_i,\bar{\mathbf{y}}_i) \}_{i=1}^{N}$.
Appendix~\ref{sec:consensus_estimation} provides more details on consensus estimation.

\subsection{Training of Human-AI cooperative model}
\label{sec:model_training}
To enable complementing unseen users given sparse, noisy multi-rater data, \projname{}'s training (Fig.\ref{fig:framework}) proceeds as follows: 
First, a set of representative annotator profiles are identified, each capturing distinct noisy labeling patterns among users. 
Next, noisy labels are augmented for each profile, to mitigate data sparsity. 
Finally, \modelname{} instances are trained for each profile using the augmented data, enabling cooperation and effective generalization to unseen users who can be matched to a profile.
We explain each step in more details below.

\subsubsection{Identifying Annotator Profiles}
In multi-rater datasets, annotators often exhibit similar labeling error patterns, reflecting individual biases or areas of confusion \cite{song2022learning, nl_unicon}.
We define these characteristic error patterns as \textit{annotator profiles}. For example, on CIFAR-10, an annotator profile might be characterized by frequent misclassifications of horses as deer, birds as planes, and trucks as automobiles \cite{nl_unicon}.

To identify representative annotator profiles, we first construct a fixed-length \textit{label vector} $\mathbf{s}_j$ for each annotator $j \in \mathcal{A}$. This vector represents the annotator's characteristic labeling pattern across all $C$ classes. The construction of $\mathbf{s}_j$ proceeds as follows: 
1) For each class $c \in \{1, ..., C\}$, we gather all noisy labels $\tilde{\mathbf{y}}_{i,j}$ provided by annotator $j$ for samples with consensus label $c$. This forms a set of labels for each class: $\mathcal{S}^{(c)}_j = \{ \tilde{\mathbf{y}}_{i,j} | (\mathbf{x}_i,\tilde{\mathbf{y}}_{i,j}) \in \tilde{\mathcal{D}}
\}$.
2) From each set $\mathcal{S}^{(c)}_j$, we randomly select $L$ labels.  If an annotator has provided fewer than $L$ labels for a given class, that annotator is excluded from further analysis.  The value of $L$ is chosen to balance profile representation with the inclusion of as many annotators as possible; in our experiments, $L$ is set to 20.
3) The selected noisy labels are then concatenated across all $C$ classes to form the label vector $\mathbf{s}_j$ where each $l_i^{(c)}$ is a noisy label $\tilde{\mathbf{y}}_{i,j}$ selected from $\mathcal{S}^{(c)}_j$.
\begin{equation}
    \mathbf{s}_j = [\underbrace{l^{(1)}_1, ..., l^{(1)}_L}_{\text{Class 1}}, \underbrace{l^{(2)}_1, ..., l^{(2)}_L}_{\text{Class 2}}, ..., \underbrace{l^{(C)}_1, ..., l^{(C)}_L}_{\text{Class C}}]
\label{eq:clustering_vector}
\end{equation}

This process yields a set of label vectors, $\mathcal{L} = \{ \mathbf{s}_j \}_{j \in \mathcal{A}}$. Although the specific labels within each $\mathbf{s}_j$ will vary, the consistent class ordering allows for direct comparison of labeling patterns across annotators.

To identify distinct annotator profiles, we cluster the label vectors in $\mathcal{L}$ using Fuzzy K-Means, chosen for its robustness to noise in the label vectors \cite{fuzKmean_robust, dehariya2010clustering}. The number of clusters, $K$, is determined using the silhouette score (see Appendix~\ref{sec:silhouette}), which provides a measure of cluster cohesion and separation. Each annotator is then assigned to the profile (cluster) to which their label vector has the highest membership score.

\subsubsection{Noisy-label Augmentation}
Because each annotator labels only a subset of the data, and annotators are grouped into profiles, the original training set $\tilde{\mathcal{D}}$ is now divided into profile-specific subsets $\{\tilde{\mathcal{D}}_k\}_{k=1}^K$ (where $\tilde{\mathcal{D}}_k$ contains labels from users in profile $k$) resulting in sparse data for training each profile-specific model.
To address this data sparsity while preserving profile-specific noise patterns, we perform noisy label augmentation on each profile, enabling the model to learn these patterns during training.

This augmentation process leverages the characteristic noisy labeling patterns captured by each profile.  Specifically, for each profile $k$, we estimate a \textit{label transition matrix} $\mathbf{T}_k \in [0,1]^{C \times C}$. This matrix quantifies the probability of annotators within profile $k$ assigning a noisy label $n$ to a sample having consensus label $c$, effectively capturing the profile specific biases. The transition matrix is estimated as,
\begin{equation}
    \mathbf{T}_{k}(c,n) = \frac{1}{|\mathcal{S}^{(c)}_k|} \sum_{\tilde{\mathbf{y}}_{i} \in \mathcal{S}^{(c)}_k} \mathbb{I}[\tilde{\mathbf{y}}_{i} = n]
\label{eq:transition_matrix}
\end{equation}
where $\mathcal{S}^{(c)}_k = \bigcup_{j \in \mathcal{A}_k} \mathcal{S}^{(c)}_j$ represents the set of noisy labels provided by annotators in profile $k$ ($\mathcal{A}_k \subset \mathcal{A}$) for samples with consensus label $c$, and $\mathbb{I}[\cdot]$ is the indicator function.

New noisy labels are then generated for each profile by sampling from this profile-specific transition matrix $\mathbf{T}_k$. This augmentation process ensures sufficient training data for each profile-specific model while preserving the characteristic noise patterns of that profile, allowing the model to learn to effectively complement the users within that profile.

Each element $\mathbf{T}_{k}(c, n)$ of the transition matrix (Eq. \ref{eq:transition_matrix}) represents the probability, $P(\tilde{Y} = n | \bar{Y} = c, R = k)$, that an annotator in profile $k$ ($R=k$) assigns the noisy label $\tilde{Y} = n$ to a sample with consensus label $\bar{Y} = c$.  To augment the data for profile $k$, we proceed as follows: For each data sample $\mathbf{x}_i$ in the profile's subset $\tilde{\mathcal{D}}_k$, we retrieve its consensus label $c$ from $\bar{\mathcal{D}}$.  Using the $c$-th row of the transition matrix $\mathbf{T}_k$, which represents the categorical distribution over noisy labels for class $c$, we sample $G$ new noisy labels, $\{\hat{\mathbf{y}}_{i,g}\}_{g=1}^{G}$. This results in an augmented training set for profile $k$: $\hat{\mathcal{D}}_k = \{ (\mathbf{x}_i, \{\hat{\mathbf{y}}_{i,g}\}_{g=1}^{G} )\}_{i=1}^{N}$.

\subsubsection{Training Human-AI Cooperative Model}
To effectively complement unseen users by leveraging learned annotator profiles and their associated augmented noisy labels, we introduce the \modelname{} architecture (final step of Training, Fig. \ref{fig:framework}). The model consists of three components. 
1) A base model, $f_{\psi_{k}}:\mathcal{X} \to \mathbb{R}^C$, extracts features from the input data, transforming it into a logit vector. 
2) A human label encoder, $h_{\phi_{k}}:\mathcal{Y} \to \mathbb{R}^C$, models the profile-specific noisy labeling patterns.
Finally, 3) a decision model, $d_{\zeta_{k}}:\mathbb{R}^C \times \mathbb{R}^C \to \Delta^{C-1}$ learns the joint noise distribution of $f_{\psi_{k}}$ and $h_{\phi_{k}}$ and produces a categorical distribution across classes  where we use the one with highest probability as the final prediction.
The whole model $m_{\theta_{k}}:\mathcal{X} \times \mathcal{Y} \to \Delta^{C-1}$ is defined as:
\begin{equation}
    m_{\theta_{k}}(\mathbf{x},\hat{\mathbf{y}}) = d_{\zeta_{k}}( f_{\psi_{k}}(\mathbf{x})  \oplus h_{\phi_{k}}(\hat{\mathbf{y}}) ),
    \label{eq:decision_model}
\end{equation}
where $\theta_{k} = \{ \psi_{k},\phi_{k},\zeta_{k} \}$, and $\oplus$ represents the concatenation operator.
Note that, for each profile $k$, we train a single $m_{\theta_{k}}$ using profile's $\hat{\mathcal{D}}_k$.
The base model $f_{\psi_{k}}(.)$ could use any architecture.
Similarly, $h_{\phi_{k}}(.)$ and $d_{\zeta_{k}}(.)$ can be of different architectures; we configured them as a two-layer and three-layer multi-layer perceptron, respectively, with ReLU activations. The model in eq.\ref{eq:decision_model} is trained with:
\begin{equation}
\scalebox{0.9}{
$\begin{aligned}
    \{\theta_{k}^*\}_{k=1}^{K} = & \arg\min_{\{\theta_{k}\}_{k=1}^{K}} \frac{1}{K \times |\hat{\mathcal{D}}_k| \times G} \times \\
    & \sum_{k=1}^{K}  \sum_{\left(\mathbf{x}_i,\{\hat{\mathbf{y}}_{i,g}\}_{g=1}^{G}\right) \in \hat{\mathcal{D}}_k}  \ell\left(\bar{\mathbf{y}}_i,m_{\theta_{k}}(\mathbf{x}_i,\hat{\mathbf{y}}_{i,g})\right) + \\
    & \lambda  \times \ell\left(\hat{\mathbf{y}}_{i,g},\left( \mathbf{T}_k \right )^{\top} \times m_{\theta_{k}}(\mathbf{x}_i,\hat{\mathbf{y}}_{i,g})\right),
\end{aligned}$
}
\label{eq:model_optimisation}
\end{equation}
where $\bar{\mathbf{y}}_i$ is the consensus label from $\bar{\mathcal{D}}$, $\ell(.)$ is the cross-entropy loss, $\lambda\in [0, \infty]$ is a hyper-parameter, and the second loss term
is motivated by the forward correction procedure proposed by \cite{lossTerm}, transforming the clean label prediction from $m_{\theta_{k}}(.)$ into the noisy ones in $\hat{\mathcal{D}}_k$.

\subsection{User Profiling and Inference}
\label{sec:onboarding}
At test time, an unseen user undergoes a \textit{user profiling} process (User Profiling phase of Fig. \ref{fig:framework}) to determine the most matching \modelname{} instance, $m_{\theta_k}$, for classification. This process consists of two key steps:
1) Assigning the user to one of the $K$ annotator profiles identified during training.
2) Evaluating an entry condition to determine whether the cooperative model instance $m_{\theta_k}(\cdot)$ from the assigned profile should be used.

In the first step, the test user annotates a small validation set $\mathcal{V} = \{(\mathbf{x}_i,\mathbf{y}_i) \}_{i=1}^{M \times C}$.
This validation set contains $M$ samples from each $C$ class, does not overlap with the training or testing sets, and has clean labels.
A key advantage of \projname{} is that it only needs $M \times C$ samples to adapt to unseen users—unlike prior methods that require extensive expert annotations to train additional behavioral models.
From the obtained labels from the test user, we construct a label vector having the same format as Eq. \ref{eq:clustering_vector}
and classify the user into a profile identified during training phase using a one-versus-all (OVA) support vector machine (SVM) classifier. OVA SVM takes the test user's label vector and outputs a categorical distribution across profiles where we choose the one with highest prediction as the matching profile.
This empirically leads to better performance compared to soft assignment (see Ablation Sec.\ref{sec:ablation}).

The OVA SVM classifier is trained on label vectors collected from users belong to identified profiles during the training. Specifically, 
user labels for $M$ samples from each of the $C$ classes (using consensus labels to determine the class), are randomly collected, formatted as in (Eq. \ref{eq:clustering_vector}), labeled with their corresponding profile, and used for training.
Given their strong performance in high dimensions with limited data, SVMs are well-suited for this task.

In the second step, building on ~\cite{Bayesian_Steyvers22}, the entry condition compares the accuracy of the base model $f_{\psi_{k}}(.)$ and the testing user on the validation set $\mathcal{V}$.
The $m_{\theta_{k}}(.)$ from the predicted profile is paired with the test user if the base model outperforms the user; else, the user is rejected.
Once the test user is paired with a model, they perform cooperative classification (Inference of Fig.\ref{fig:framework}). $m_{\theta_{k}}(.)$ is evaluated on a test set $\mathcal{T} = \{ (\mathbf{x}_i,\mathbf{y}_i) \}$ having clean labels that does not overlap with training or validation sets.

\subsection{Measures for H-AI Cooperative Classification}
\label{sec:new_measures}
As part of cooperative classification, \projname{} alters the user labels to improve accuracy. These alterations can be positive (correcting an incorrect user label) or negative (changing a correct user label to an incorrect one). To capture this, we introduce positive and negative alteration measures,
\begin{equation}
\scalebox{0.88}{$
    \begin{aligned}
    \parbox{1.4cm}{Positive\\Alteration\\$A_{+}$} = \frac{1}{ |\mathcal{A}|} \sum_{j=1}^{|\mathcal{A}|} 
    \frac{|\mathcal{M^\mathbf{c}}_j|}{|\mathcal{I}_j|}
    \hspace{0.4cm}
    \parbox{1.4cm}{Negative\\Alteration\\$A_{-}$} = \frac{1}{ |\mathcal{A}| } \sum_{j=1}^{|\mathcal{A}|} 
    \frac{|\mathcal{M^\mathbf{e}}_j|}{|\mathcal{R}_j|}
    \end{aligned}
    $}
\label{eq:alt}
\end{equation}
where $\mathcal{I}_j  = \{ i | i \in \mathcal{T}  \text{ and } \tilde{\mathbf{y}}_{i,j} \neq \mathbf{y}_i  \}$ represents the set of samples incorrectly labeled by the $j^{th}$ user,
$\mathcal{M^\mathbf{c}}_j = \{ i | i \in \mathcal{I}_j \text{ and } \ddot{\mathbf{y}}_{i,j} = \mathbf{y}_i  \}$ represents the set of samples labeled incorrectly by user $j$ but corrected by the model, 
$\mathcal{R}_j = \{ i | i \in \mathcal{T}  \text{ and } \tilde{\mathbf{y}}_{i,j} = \mathbf{y}_i  \}$ represents the set of samples correctly labeled by the $j^{th}$ user,
$\mathcal{M^\mathbf{e}}_j = \{ i | i \in \mathcal{R}_j \text{ and } \ddot{\mathbf{y}}_{i,j} \neq \mathbf{y}_i \}$ represents the set of samples labeled correctly by user $j$ but later mislabeled by the model.
Note that $\mathbf{y}_i$ represents the test set clean label and $\ddot{\mathbf{y}}_{i,j} = \mathsf{Scalar}(m_{\theta_{k}}(\mathbf{x}_i,\tilde{\mathbf{y}}_{i,j}))$, with the function 
$\mathsf{Scalar}
$ 
returning a scalar label representing the class with the largest prediction from the model $m_{\theta_{k}}(.)$. 
In eq.\ref{eq:alt}, $A_{+}$ measures the proportion of a user's incorrect labels that the model successfully corrected.
In contrast, $A_{-}$, in eq.\ref{eq:alt}, measures the proportion of a user's correct labels that were incorrectly altered by the model.
In edge cases where the annotator is perfect or always wrong, $A_+$ and $A_-$ becomes zero respectively, for division by zero.

We also assess original accuracy (before labels alterations) and post-alteration accuracy (after \modelname{} alters labels). These are computed per user and averaged across all users to determine overall improvement.
An effective model should have high $A_{+}$, high post-alteration accuracy and low $A_{-}$.

\section{Experiment Setup}
\label{sec:experimental_setup}
\subsection{Datasets} 
\textbf{CIFAR-10} \cite{c10} includes 50000 training, 200 validation, and 9800 testing images, across 10 classes.
\textbf{CIFAR-10N} \cite{c10n} extends CIFAR-10's training set with labels collected from 747 annotators, with each image having three independent labels.
\textbf{CIFAR-10H} \cite{c10h} expands CIFAR-10's testing set with labels from 2571 annotators, resulting in an average of 51 labels per image.
\textbf{Fashion-MNIST-H} \cite{fmnisth} extends Fashion-MNIST's \cite{fmnist} testing set with labels from 885 annotators, averaging 66 labels per image. We use this testing set as the training set, with 200 images from the original training set allocated for validation and the remainder for testing.
Lastly, \textbf{Chaoyang} \cite{chaoyang} is a four-class pathological dataset with 4021 training, 80 validation, and 2059 testing images, each having three expert labels in the training set.
Unless stated otherwise, above validation sets are composed of randomly selected subsets of samples from the respective original test sets, with selected samples excluded from testing.

\subsection{Setup on datasets with simulated annotators}
On CIFAR-10, a pairwise flipping experiment is conducted where 8 out of 10 classes have clean labels, but in two classes, 60\% of samples have labels flipped. Three user profiles are simulated, one that flips labels between classes airplane$\leftrightarrow$bird, another profile that flips horse$\leftrightarrow$deer, and the other flips truck$\leftrightarrow$automobile.
For each profile, five training and five testing users are simulated, resulting in 15 unique users in each set.
Labels from training users combined with training samples, form $\tilde{\mathcal{D}}$ from which we identify $K$ profiles, train $m_{\theta_{k}}$ in eq. \ref{eq:decision_model} for each $k$ with a ResNet-18 \cite{resnets} as $f_{\psi_{k}}(.)$, and train OVA SVM.

\subsection{Setup on datasets with real annotators}
With CIFAR-10N, we conduct two experiments. In the first experiment, the labels from 747 annotators form $\tilde{\mathcal{D}}$. Of these, 159 annotators who labelled at least 20 images per class ($L=20$) are selected, split into 79 training users and 80 testing users. The training users' labels are used to identify $K$ profiles where $K$ is chosen from silhouette score, train $m_{\theta_{k}}$ for each profile, and train the OVA SVM. During testing, noisy-label transition matrices are estimated using annotator labels and consensus labels for each test user, resulting in 80 noisy test sets.
In the second CIFAR-10N experiment, users in CIFAR-10H are used as a testing set. Noise transition matrices are estimated and used to simulate noisy annotations for each testing user, resulting in unique noisy test sets for all 2571 users. 
For Fashion-MNIST-H, labels from 885 annotators form $\tilde{\mathcal{D}}$. 366 annotators who labeled at least 20 images per class are selected, split into 183 training and 183 testing sets. Training users are used to identify $K$ profiles, train $m_{\theta_{k}}$, and train an OVA SVM.
During testing noisy-label transition matrices are estimated for each testing user and produced 183 noisy testing sets.
Chaoyang dataset has three annotators per image, forming $\tilde{\mathcal{D}}$ which is used to build $K$ profiles, train $m_{\theta_{k}}$, and train an OVA SVM. During testing, noisy-label transition matrices are estimated, resulting in three noisy test users.
More details on experiment setup, data preparation, and implementation are in Appendix \ref{supp:experimental_setup}.

\begin{table}[tb]
\centering
\small
\caption{Accuracy of \projname{} vs. related methods. Unlike others, \projname{} outperforms without training ground truths and by complementing unseen users.
(Missing values due to lack of sparse multi-rater support.)
}
\resizebox{1\linewidth}{!}{%
\begin{tabular}{lccccc} 
\hline
\multirow{2}{*}{Method} &  \multicolumn{1}{c}{CIFAR} & \multicolumn{1}{c}{CIFAR} & \multicolumn{1}{c}{CIFAR} & \multicolumn{1}{c}{F-MNI-} & \multicolumn{1}{c}{Chao-} \\ 
                        & -10 & -10N   & -10H   & ST-H     &    yang   \\
\hline\hline
\multicolumn{6}{c}{Learning To Defer} \\ 
\hline
MOE \cite{predictResponsibly_madras} & - & 0.831 & 0.812 & 0.600 & 0.583 \\
CC \cite{raghu19_triage} & - & 0.970 & 0.971 & 0.801 & 0.863 \\
CE \cite{consistentest_Mozannar2020} & - & 0.949 & 0.967 & 0.729 & 0.706 \\
DifT \cite{okati21DiffTriage} & - & 0.940 & 0.944 & 0.704 & 0.765 \\
OVA \cite{verma22OVA} & - & 0.959 & 0.974 & 0.794 & 0.845 \\
WSP \cite{whoshould_mozannar23} & - & 0.948 & 0.976 & 0.775 & 0.872 \\
\hline
\multicolumn{6}{c}{Learning to Defer to Unseen Users} \\ 
\hline
L2D-Pop \cite{l2d_2pop} & 0.947 & - & - & - & 0.970 \\
EA-L2D \cite{ea_l2d} & 0.820 & - & - & - & - \\ 
\hline
\multicolumn{6}{c}{Learning To Complement} \\ 
\hline
LECOMH \cite{l2cmh} & - & - & 0.988 & - & 0.988 \\
LECODU \cite{l2cdmu} & 0.951 & - & 0.989 & - & 0.990 \\ 
\hline
\multirow{2}{*}{\textbf{L2CU (Ours)}} & \textbf{0.968} & \textbf{0.989} & \textbf{0.993} & \textbf{0.878} & \textbf{0.991} \\ 
                                             & $\pm$\textbf{0.002}
                                             & $\pm$\textbf{0.001}
                                             & $\pm$\textbf{0.002}
                                             & $\pm$\textbf{0.008}
                                             & $\pm$\textbf{0.004} \\ 
\hline
\end{tabular}
}
\label{table:benchOthers}
\end{table}

\begin{table*}[tb]
\centering
\begin{minipage}{0.58\linewidth}
\centering
\caption{Number of users who improved (I), maintained (M), or did not improve (NI) out of the test users profiled and satisfied entry condition. Includes original vs. post-alteration accuracy with positive ($A_+$) and negative ($A_-$) alterations along with chosen $K$ from silhouette score.}
\resizebox{\linewidth}{!}{
\small
\begin{tabular}{lc|cccc|cc|cc} 
\hline
Dataset & \begin{tabular}[c]{@{}c@{}}$K$ (Silhou-\\ette score)\end{tabular} & \begin{tabular}[c]{@{}c@{}}Test\\Users\end{tabular} & I & M & NI & \begin{tabular}[c]{@{}c@{}}Original\\Accuracy\end{tabular} & \begin{tabular}[c]{@{}c@{}}Post. Alt.\\Accuracy\end{tabular} & $A_+$ & $A_-$ \\ 
\hline\hline
\multicolumn{10}{c}{With simulated annotators} \\ 
\hline
CIFAR-10 & 3 (0.34) & 15 & 15 & 0 & 0 & 0.880 & 0.968 & 0.953 & 0.09 \\
\hline
\multicolumn{10}{c}{With real annotators} \\ 
\hline
CIFAR-10N & 2 (0.01) & 80 & 80 & 0 & 0 & 0.836 & 0.989 & 0.954 & 0.004 \\
CIFAR-10H & 2 (0.01) & 2022 & 2022 & 0 & 0 & 0.940 & 0.993 & 0.939 & 0.004 \\
F-MNIST-H & 2 (0.09) & 182 & 182 & 0 & 0 & 0.662 & 0.878 & 0.758 & 0.073 \\
Chaoyang & 3 (0.99) & 2 & 2 & 0 & 0 & 0.858 & 0.988 & 0.968 & 0.045 \\
\hline
\end{tabular}}
\vspace{0.8em}
\label{table:ex_users&acc}
\end{minipage}
\hfill
\vspace{-0.1em}
\begin{minipage}{0.40\linewidth}
\centering
\includegraphics[width=0.82\linewidth]{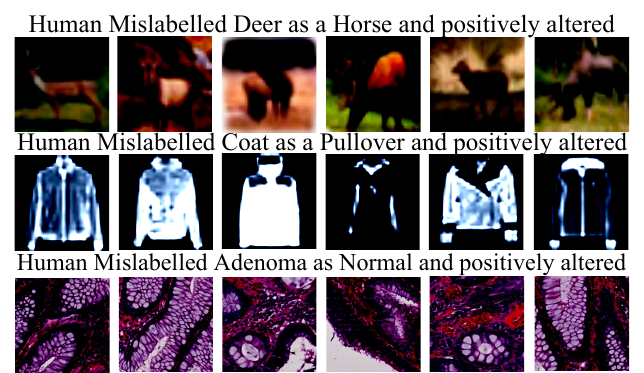}
\captionof{figure}{Positive alterations made by the \projname{} on CIFAR-10N, Fashion-MNIST-H and Chaoyang
experiments (top to bottom). 
}
\label{fig:alt_vis}
\end{minipage}
\end{table*}

\subsection{Training details} 
Data augmentation policies in \cite{cifar_AutoAug, fmnist_randaug} were adopted for CIFAR-10 and Fashion-MNIST respectively while Chaoyang is limited to random resized crops of $224\times224$.
Training runs for 500 epochs with early stopping after 20 unimproved epochs.
As found empirically (see Ablation Sec.\ref{sec:ablation}), we keep $\lambda=0.1$ that yields optimal performance.
We use Imagenet1K \cite{imgnet} pre-trained backbone models as base model.
Adam and NAdam optimize $f_{\psi_k}(.)$ and $m_{\theta_k}(.)$ respectively, in PyTorch on a NVIDIA RTX 4090.

\section{Results}
Results are reported for unseen test users who were profiled and met the entry condition.
Table~\ref{table:benchOthers} compares \projname{} with competing methods and 
Table~\ref{table:ex_users&acc} presents the post-alteration accuracy relative to the users' original accuracy, with positive and negative alterations from eq.\ref{eq:alt} for $K$ selected from silhouette score and Fig.\ref{fig:alt_vis} with sample positive alterations.

\label{sec:ev_results}
\subsection{Comparison with related methods}\label{sec:comparative_analysis}
Table~\ref{table:benchOthers} compares \projname{} with competing methods from L2D, L2D with unseen users, and L2C on both simulated (CIFAR-10) and real annotator datasets (CIFAR-10N, CIFAR-10H, Fashion-MNIST-H, Chaoyang).
For a fair comparison, Chaoyang results are reported for all users without applying the entry condition.
We prioritize comparisons with real annotator datasets and use simulations only when necessary.
Some values are missing due to requiring the same users in training and testing sets, not supporting multi-rater settings with sparse labels (where we resort to simulation data), or unavailable source code (where we report values from original work).
Unlike literature methods trained with ground truth (except LECODU), our models are trained without it (with consensus) yet still outperform them.
Low standard deviation indicate steady improvements across users and datasets.

\subsection{Results of datasets with simulated annotators}
The first row of Table~\ref{table:ex_users&acc} show the number of testing users who improved (I), maintained (M), or did not improved (NI) with \projname{} in the CIFAR-10 simulation.
The comparison between original and post-alteration accuracy reveals that all 15 testing users improved, with post-alteration accuracy exceeding the original.  
The last two columns show a large $A_{+}$ contrasted with a low $A_{-}$, highlighting a high proportion of positive alterations.

\subsection{Results of datasets with real annotators}
Table~\ref{table:ex_users&acc} shows that all test users improved their post-alteration accuracy with \projname{} by approximately 18\%, 5\%, 32\%, and 15\% for CIFAR-10N, CIFAR-10H, Fashion-MNIST-H, and Chaoyang, respectively along with a high positive alteration rate compared to negative alterations.
Interestingly, real annotator datasets (except Chaoyang) have low silhouette scores likely because they include many annotators, each introducing subtle noise patterns, making profiles harder to distinguish. Being a real annotator dataset, with fewer annotators, Chaoyang  has the highest score.
However, \projname{} manages to improve users in such challenging conditions.

\subsection{Adapting to the Text Domain}
Table \ref{table:agnews} evaluates \projname{} in the text domain using AgNews \cite{agnews} dataset following a simulation setup similar to CIFAR-10.
All 15 test users improved post-alteration accuracy over original and a high positive alteration proportion, promising that adapting to the text domain is a possibility.
See appendix \ref{supp:experimental_setup} for experiment setup.
\begin{table}[h]
\small
\centering
\caption{\projname{} performance on Text domain with Agnews}
\resizebox{\linewidth}{!}{%
\begin{tabular}{c|cccc|cc|cc}
\hline
\begin{tabular}[c]{@{}c@{}}$K$ (Silhouette\\score)\end{tabular} & \begin{tabular}[c]{@{}c@{}}Test\\Users\end{tabular} & I & M & NI & \begin{tabular}[c]{@{}c@{}}Original\\Accuracy\end{tabular} & \begin{tabular}[c]{@{}c@{}}Post. Alt.\\Accuracy\end{tabular} & $A_+$ & $A_-$ \\ 
\hline\hline
3 (0.44) & 15 & 15 & 0 & 0 & 0.700 & 0.980 & 0.975 & 0.016 \\
\hline
\end{tabular}
}
\label{table:agnews}
\end{table}

\subsection{Role of Annotator Profiles}
\label{sec:role_of_clus}
To evaluate the impact of annotator profiles on unseen user performance, we compare results with and without profiles (Tables \ref{table:ex_users&acc} and \ref{table:cluster_role}). 
The absence of profiles leads to a higher number of users whose performance is not improved (NI), lower post alteration accuracy, and lower positive label alterations across all datasets compared to \textit{with} profiles.
This demonstrates that profiles are crucial for improving the performance of unseen users.
\begin{table}
\small
\centering
\caption{Results without annotator profiles}
\resizebox{\linewidth}{!}{%
\begin{tabular}{lccc|cc|cc} 
\hline
Dataset & I & M & NI & \begin{tabular}[c]{@{}c@{}}Original\\Accuracy\end{tabular} & \begin{tabular}[c]{@{}c@{}}Post. Alt.\\Accuracy\end{tabular} & $A_+$ & $A_-$ \\ 
\hline\hline
CIFAR-10 & 5 & 0 & 10 & 0.880 & 0.835 & 0.815 & 0.261 \\
\hline
CIFAR-10N & 69 & 0 & 11 & 0.836 & 0.918 & 0.893 & 0.006 \\
CIFAR-10H & 1949 & 0 & 73 & 0.940 & 0.911 & 0.882 & 0.006 \\
F-MNIST-H & 166 & 2 & 14 & 0.662 & 0.854 & 0.635 & 0.181 \\
Chaoyang & 1 & 0 & 1 & 0.858 & 0.915 & 0.704 & 0.065 \\ 
\hline
\end{tabular}}
\label{table:cluster_role}
\end{table}

To validate the silhouette score's selection of the optimal number of clusters ($K$), we visualize the profiles identified in the CIFAR-10 simulation experiment with $K=3$ (Fig. \ref{fig:n_patterns_c10}, Appendix \ref{supp:vis_noise_profiles}). 
The estimated noise matrices for these profiles closely resemble the ground-truth noise patterns used in the simulation: one profile exhibits noise between airplane and bird, another between horse and deer, and a third between truck and automobile. This confirms both the effectiveness of the silhouette score in determining $K$ and the 
ability of clustering approach to identify distinct user noise patterns.

\subsection{Distribution of Joint Decisions}
Table~\ref{table:ex_choices} shows decision distributions for unseen test users in real annotator experiments, comparing humans, the base model, and their joint cooperation.
Each decision—by the human, base model $f_{\psi_{k}}(.)$, or cooperation $m_{\theta_{k}}(.)$—is marked as correct (\checkmark) if it matches the test set target label or incorrect (\ding{55}) otherwise.

\begin{table}[h]
\small
\centering
\caption{Proportion that each combination of Human, AI, or Cooperation is correct \checkmark or incorrect \ding{55}; Columns sum to 1.
}
\resizebox{\linewidth}{!}{%
\begin{tabular}{ccc|cccc} 
\hline
\multirow{2}{*}{Human} & \multirow{2}{*}{\begin{tabular}[c]{@{}c@{}}AI\\$f_{\psi_{k}}(.)$\end{tabular}} & \multirow{2}{*}{\begin{tabular}[c]{@{}c@{}}Coopera-\\tion $m_{\theta_{k}}(.)$\end{tabular}} 
& \multirow{2}{*}{\begin{tabular}[c]{@{}c@{}}CIFAR\\-10N \%\end{tabular}} & \multirow{2}{*}{\begin{tabular}[c]{@{}c@{}}CIFAR\\-10H \%\end{tabular}} & \multirow{2}{*}{\begin{tabular}[c]{@{}c@{}}F-MNI-\\ST-H \%\end{tabular}} & \multirow{2}{*}{\begin{tabular}[c]{@{}c@{}}Chaoy-\\ang \%\end{tabular}} \\ 
 &  &  &  &  &  & \\ 
\hline\hline
\ding{55} & \checkmark & \checkmark  & 05.15   & 05.59 & 04.47  & 03.35 \\
\checkmark & \ding{55} & \checkmark  & 00.65   & 02.26 & 15.05  & 01.82 \\
\checkmark & \checkmark & \checkmark & 93.79   & 91.35 & 72.13  & 92.16 \\
\ding{55} & \ding{55} & \checkmark   & 00.05   & 00.05 & 04.29  & 00.13 \\ 
\hline
\ding{55} & \checkmark & \ding{55}  & 00.13 & 00.19 & 00.33 & 00.49 \\
\checkmark & \ding{55} & \ding{55}  & 00.11 & 00.39 & 01.38 & 01.29 \\
\checkmark & \checkmark & \ding{55} & 00.00 & 00.00 & 00.20 & 00.00 \\
\ding{55} & \ding{55} & \ding{55}   & 00.12 & 00.17 & 02.17 & 00.76 \\
\hline
\end{tabular}
}
\label{table:ex_choices}
\end{table}

\begin{table*}[t]
\begin{minipage}[t]{0.24\textwidth}
    \centering
    \caption{Performance vs. importance of components. Checkmarks indicate (not-)having component\textbf{---Ablation 1}}
    \resizebox{\textwidth}{!}{%
\centering
\small
\resizebox{\linewidth}{!}{%
\begin{tabular}{cc|ccc} 
\hline
\begin{tabular}[c]{@{}c@{}}$h_{\phi}(.)$\\ \end{tabular} & \begin{tabular}[c]{@{}c@{}}$d_{\zeta}(.)$\\ \end{tabular} & \begin{tabular}[c]{@{}c@{}}Post Al-\\t. Acc\end{tabular} & $A_+$ & $A_-$ \\ 
\hline\hline
\ding{55} & \ding{55} & 0.705 & 0.007 & 0.159 \\
\checkmark & \ding{55} & 0.774 & 0.043 & 0.083 \\
\ding{55} & \checkmark & 0.861 & 0.833 & 0.134 \\
\checkmark & \checkmark & 0.989 & 0.954 & 0.004 \\
\hline
\end{tabular}
}
}
    \label{table:abl_model_parts}
\end{minipage}
\hfill
\begin{minipage}[t]{0.22\textwidth}
    \centering
    \caption{Performance vs. noisy label augmentation hyper-parameter $G$ \textbf{---Ablation 2}}
    \resizebox{\textwidth}{!}{%

\small
\resizebox{0.6\linewidth}{!}{%
\begin{tabular}{cccc} 
\hline
\multirow{2}{*}{$G$} & \multirow{2}{*}{\begin{tabular}[c]{@{}c@{}}Post Alt.\\Acc.\end{tabular}} & \multirow{2}{*}{$A_+$} & \multirow{2}{*}{$A_-$} \\ 
 & & & \\
\hline\hline
0 & 0.615 & 0.411 & 0.302  \\
1 & 0.980 & 0.953 & 0.004  \\
3 & 0.983 & 0.954 & 0.004  \\
5 & 0.989 &  0.952 & 0.004 \\
\hline
\end{tabular}
}
}
    \label{table:abl_aug}
\end{minipage}
\hfill
\begin{minipage}[t]{0.23\textwidth}
    \caption{Performance vs. noise rate \textbf{---Ablation 3}}
    \vspace{0.45cm}
    \resizebox{\textwidth}{!}{%
\small
\centering
\resizebox{\linewidth}{!}{%
\begin{tabular}{cccc} 
\hline
\begin{tabular}[c]{@{}c@{}}Noise\\Rate\end{tabular} & \begin{tabular}[c]{@{}c@{}}Post Alt.~\\Acc.\end{tabular} & $A_+$ & $A_-$ \\ 
\hline\hline
40\% & 0.992 & 0.973 & 0.001 \\
60\% & 0.968 & 0.953 & 0.088 \\
80\% & 0.879 & 0.944 & 0.134 \\
90\% & 0.868 & 0.875 & 0.195 \\
\hline
\end{tabular}
}
}
    \label{table:abl_noiserates}
\end{minipage}
\hfill
\begin{minipage}[t]{0.24\textwidth}
    \centering
    \caption{Performance vs. different backbones as base model $f_{\psi_{k}}(.)$ \textbf{---Ablation 4}}
    \vspace{0.3cm}
    \resizebox{\textwidth}{!}{%
    \begin{tabular}{lcccc} 
\hline
\begin{tabular}[c]{@{}c@{}}Backbone\\Model\end{tabular} & \begin{tabular}[c]{@{}c@{}}Post Alt.\\Acc.\end{tabular} & $A_+$ & $A_-$ \\ 
\hline\hline
ResNet-50 & 0.968 & 0.862 & 0.013 \\
\begin{tabular}[c]{@{}l@{}}DenseNet-\\121\end{tabular} & 0.969 & 0.854 & 0.011 \\
Vit/B-16 & 0.989 & 0.954 & 0.004 \\ 
\hline
\end{tabular}
    }
    \label{table:abl_bases_acc}
\end{minipage}
\end{table*}

According to Table~\ref{table:ex_choices}, the majority of correct joint decisions and the lowest proportion of incorrect joint decisions occurs when both human and AI predictions are correct, as expected from a cooperation.
In addition, joint decisions tend to be correct when at least one party is correct, showing the effectiveness of cooperation.
Interestingly, there are cases where the cooperative decision is correct when both individual parties are wrong which we discuss this in Sec.\ref{sec:discussion}.

\section{Ablation Studies}\label{sec:ablation}
\textbf{(Ablation 1) Performance vs. Model Components:}
Table \ref{table:abl_model_parts} evaluates the importance of Human Label Encoder $h_\phi$ and Decision Model $d_\zeta$ for overall performance by turning each off separately and together with CIFAR-10N.
By changing the eq. \ref{eq:decision_model}, when $h_\phi$ is off,
$m_{\theta_{k}}(\mathbf{x},\hat{\mathbf{y}}) = d_{\zeta_{k}}( f_{\psi_{k}}(\mathbf{x})  \oplus \hat{\mathbf{y}})$.
When $d_\zeta(.)$ is off, 
$m_{\theta_{k}}(\mathbf{x},\hat{\mathbf{y}}) =  f_{\psi_{k}}(\mathbf{x})  \oplus h_{\phi_{k}}(\hat{\mathbf{y}})$.
When both are off,
$m_{\theta_{k}}(\mathbf{x},\hat{\mathbf{y}}) = f_{\psi_{k}}(\mathbf{x})  \oplus \hat{\mathbf{y}}$.

Accuracy becomes lowest with both components off and highest when both are active. Using only one is suboptimal, showing the importance of added components for unseen users.
\textbf{(Ablation 2) Performance vs. $G$:}  
Table \ref{table:abl_aug} extends the CIFAR-10N experiment to study post-alteration accuracy for augmentation times $G \in \{0,1,3,5\}$. Accuracy jumps significantly from $G=0$ to $G=1$ showing the effect of noisy label augmentation, with steady gains for $G>1$.
\textbf{(Ablation 3) Performance vs. Noise Rate:}  
Table \ref{table:abl_noiserates} expands the CIFAR-10 simulation (Section \ref{sec:experimental_setup}) to study asymmetric noise rates (40\%–90\%). Our approach remains robust, maintaining accuracy above 86\% across cases.
\textbf{(Ablation 4) Evaluating Backbone Models:}  
We extend the CIFAR-10N experiment with DenseNet-121, ResNet-50, and ViT/B-16 as base model. Tab.\ref{table:abl_bases_acc} shows consistent performance across all models, being agnostic to the backbone. 

We perform four additional ablations in which we keep the findings here and the results in the Appendix \ref{supp:additional_abl}.
\textbf{(Ablation 5) Performance vs. $K$:}
Furthering role of clusters in Sec. \ref{sec:role_of_clus}, we study the effect of having a $K$, different from the silhouette optimal.
Table \ref{table:abl_K} extends CIFAR-10N experiment to $K \in \{1,2,3,6,10\}$ showing that post alteration accuracy improves from $K=1$ (no clusters) to $K=2$ (optimal for CIFAR-10N), but drops for $K>2$, likely due to over-adaptation, as higher \( K \) reduces training users per profile, making the model less generalizable.
\textbf{(Ablation 6) Performance vs. $\lambda$:}
We study the impact of $\lambda$ in the loss function eq.\ref{eq:model_optimisation} on post-alteration accuracy using CIFAR-10N with ResNet-50, DenseNet-121, and Vit/B-16. Experiments with $\lambda \in \{0, 0.01, 0.1, 1, 10\}$ (Table~\ref{table:abl_lambda}) show that the highest accuracy is centered around $\lambda=0.1$ across all models.
\textbf{(Ablation 7) Performance vs. SVM Profiling Error:}
To assess the impact of SVM profiling errors, we randomly assigned test users a profile different from the one predicted by the SVM.
Table \ref{table:svm_profiling_mistakes} shows the expected surge in not improved (NI) users and the drop in post alteration accuracy.
This performance drop contrasts with Table~\ref{table:benchOthers}, where correct SVM assignments yield higher post-alteration accuracy.
\textbf{(Ablation 8) Performance vs. Profile Assignment Method:}
Hard assignment uses the prediction from the most matching profile model $m_{\theta_k}$ at inference, while soft assignment averages predictions from all $m_{\theta_{k \in {\{1,...,K\}}}}$, weighted by OVA SVM's profile probabilities. Table \ref{table:abl_soft_assignements} shows that hard assignment performs better than soft profile assignment across all datasets.
All ablation studies adopt the setup in Section \ref{sec:experimental_setup} and use the $K$ selected by silhouette score in Table \ref{table:ex_users&acc}.

\section{Discussion}\label{sec:discussion}
\subsection{Can cooperation correct joint mistakes?}
Table \ref{table:ex_choices} reveals an interesting phenomenon: the cooperative decision can be correct even when both human and AI are wrong.
This happens because the decision model $d_{\zeta_{k}}(.)$ learns to exploit the joint label noise distribution of the base model and human to correct their combined errors.
A necessary condition for this is $P(C|\neg A,\neg B) > 0$, where $A$, $B$ and $C$ represent events that correct predictions from the AI, human, and human-AI cooperation, respectively. 
Given that, base model and human can make mistakes and assuming events $A$ and $B$ are independent and independent given $C$, we trivially obtain: $P(C|\neg A,\neg B) = \frac{P(\neg A,\neg B|C).P(C)}{P(\neg A,\neg B)} =\frac{(1-P(A|C)).(1-P(B|C)).P(C)}{(1-P(A))(1-P(B))}>0$ because $0 < P(B|C),P(A|C), P(A), P(B), P(C) < 1$.

\subsection{When would cooperation reduce performance?}
Although generally improves joint performance, \projname{} may degrade the performance of experts with near-perfect accuracy.
We observe this with Chaoyang dataset having one expert with 99.6\% original accuracy and 99.3\% post alt. accuracy.
This aligns with theoretical findings in \cite{Bayesian_Steyvers22} which demonstrates that an L2C system's improvement is contingent upon the model outperforming its human counterpart.
To mitigate degrades, we introduced an entry condition in our profiling process (Sec. \ref{sec:onboarding})  that compares the base model accuracy on $\mathcal{V}$, to that of the test user.

\subsection{Future directions}
First, extending \projname{} to a distribution-agnostic user setting would be a promising direction that broadens its application.
Second, modeling temporal dynamics would enable complementing the changing user behavior over time.
Third, using meta-learning to adapt AI cooperative models to profiles would lower the computational overhead.
Finally, developing a few-shot user profiling would reduce the annotation requirements, especially for datasets with many classes.

\section{Conclusion}
This paper introduced \projname{}, a novel learning to complement framework that enables complementing unseen users in human-AI cooperative classification.
Extensive evaluations across datasets (CIFAR-10N, CIFAR-10H, Fashion-MNIST-H, Chaoyang, and AgNews) demonstrate \projname{}'s leading performance without needing ground truth.
Furthermore, the proposed label augmentation method tackles data sparsity while preserving annotator bias, and 
the alteration rate metric offers insights into the model's impact on human labels.
With the model-agnostic design and ability to leverage noisy, sparse multi-user data without access to ground truth, \projname{} offers a significant step toward human-AI cooperative classification systems.

\appendices
\section{\break Consensus Label Estimation}
\label{sec:consensus_estimation}
Many multi-rater input datasets lack ground truth labels. 
To address this, \projname{} is built to function effectively without relying on them. During training, we use Crowdlab~\cite{crowdlab} to estimate a consensus label $\bar{\mathbf{y}}_i$, which approximates the true clean label $\mathbf{y}_i$.
\cite{crowdlab} works in two steps. In the first step, it estimates a consensus by majority vote $\bar{\mathbf{y}}_{i}'$ per training sample. In the second step, it trains a classifier using the initial consensus and obtains predicted class probabilities for each training example. 
Thereafter, these predicted probabilities along with the original annotations from the raters are used to estimate a better consensus, creating the following ensemble,
\begin{equation}
    \bar{\mathbf{y}}_i = \mathbf{w}_{\gamma} \times f_{\gamma}(\mathbf{x}_i) + \mathbf{w}_{1} \times \tilde{\mathbf{y}}_{i,1} + ... +  \mathbf{w}_{|\mathcal{A}|} \times \tilde{\mathbf{y}}_{i,|\mathcal{A}|},
    \label{eq:crowdEn}
\end{equation}
where $f_{\gamma}:\mathcal{X} \to \Delta^{C-1}$ is a classifier trained with the majority vote label $\bar{\mathbf{y}}_{i}'$ to output a categorical distribution for $C$ classes, and the weights $\mathbf{w}_{\gamma},\mathbf{w}_{1},...,\mathbf{w}_{|\mathcal{A}|}$ are assigned according to an estimate of how trustworthy the model is, compared to each individual annotator.
The outcome of Crowdlab is a consensus labeled training set denoted by $\bar{\mathcal{D}} = \{ (\mathbf{x}_i,\bar{\mathbf{y}}_i) \}_{i=1}^{N}$.
Note that the consensus label is necessary only when the clean label $\mathbf{y}_i$ is latent. If such clean label is observed, then Crowdlab is no longer needed, and  \projname{} can be trained with $\mathcal{D} = \{ (\mathbf{x}_i,\mathbf{y}_i) \}_{i=1}^{N}$.

\section{\break Deciding the Optimal Number of Profiles}
\label{sec:silhouette}
We determine the optimal number of profiles $K$ with the silhouette score defined by,
\begin{equation}
\label{eq:silhouette}
S_k = \frac{1}{|A|} \sum_{j \in \mathcal{A}}^{} \frac{b(\mathbf{s}_j) - a(\mathbf{s}_j)}{\max\{a(\mathbf{s}_j), b(\mathbf{s}_j)\}},
\end{equation}
where $a(\mathbf{s}_j)$ denotes the sample's intra-profile distance (i.e., the average L2 distance to all other points in the same profile), 
$b(\mathbf{s}_j)$ represents the inter-profile distance (i.e., the lowest average L2 distance to all points in any other profile).
The mean silhouette score for $K$ profiles is defined by $S(K) = \frac{1}{K}\sum_{k=1}^{K}S_k$.
The optimal number of profiles for the dataset is identified by the $K$ that yields the highest silhouette score.

\section{\break Experimental Setup}
\label{supp:experimental_setup}

\begin{table*}[tb]
\begin{minipage}[t]{0.45\textwidth}
    \centering
    \caption{Performance vs. number of profiles $K$ on CIFAR-10N \textbf{---Ablation 5}}
    \resizebox{0.6\textwidth}{!}{%
\small
\resizebox{0.6\linewidth}{!}{%
\begin{tabular}{cclccl} 
\hline
\multirow{2}{*}{$K$} & \multirow{2}{*}{\begin{tabular}[c]{@{}c@{}}Post Alt.\\Accuracy.\end{tabular}} & \multirow{2}{*}{$A_+$} & \multirow{2}{*}{$A_-$} \\ 
 & & & \\
\hline\hline
K=1 & 0.918 & 0.893 & 0.006  \\
K=2 & 0.989 & 0.954 & 0.004  \\
K=3 & 0.988 & 0.954 & 0.004  \\
K=6 & 0.986 & 0.944 & 0.004  \\
K=10 & 0.972  & 0.914 & 0.004 \\
\hline
\end{tabular}
}
}
    \label{table:abl_K}

    \caption{Performance vs. $\lambda$ of the loss in eq.\ref{eq:model_optimisation} (with CIFAR-10N) \textbf{---Ablation 6}}
    \resizebox{\textwidth}{!}{%
\centering
\small
\resizebox{0.6\linewidth}{!}{%
\begin{tabular}{c|ccccc} 
\hline 
\multirow{2}{*}{\begin{tabular}[c]{@{}c@{}}Backbone\\model\end{tabular}} & \multirow{2}{*}{$\lambda=0$} & \multirow{2}{*}{$\lambda=0.01$} & \multirow{2}{*}{$\lambda=0.1$} & \multirow{2}{*}{$\lambda=1$} & \multirow{2}{*}{$\lambda=10$} \\ 
 & & & & & \\ 
\hline\hline
ResNet-50 & 0.929 & 0.944 & 0.968 & 0.939 & 0.929 \\
DenseNet-121 & 0.936 & 0.950 & 0.969 & 0.937 & 0.931 \\
ViT-B/16 & 0.982 & 0.982 & 0.989 & 0.976 & 0.969 \\
\hline
\end{tabular}
}
}
    \label{table:abl_lambda}   
\end{minipage}
\hfill
\begin{minipage}[t]{0.47\textwidth}
    \centering
    \caption{Performance vs. SVM profiling error. \textbf{---Ablation 7}}
    \vspace{0.23cm}
    \resizebox{0.665\textwidth}{!}{%
\centering
\resizebox{0.6\linewidth}{!}{%
\begin{tabular}{l|cc} \hline
Dataset   & \begin{tabular}[c]{@{}c@{}}Not Improved\\Users (NI)\end{tabular} & \begin{tabular}[c]{@{}c@{}}Post Alt.\\Accuracy\end{tabular}  \\ \hline\hline
CIFAR-10  & 11                                                          & 0.82                                                         \\
CIFAR-10N & 29                                                          & 0.92                                                         \\
CIFAR-10H & 181                                                         & 0.90                                                         \\
F-MNIST-H & 71                                                          & 0.84                                                         \\
Chaoyang  & All                                                         & 0.89                                                         \\ \hline
\end{tabular}
}
    }
    \label{table:svm_profiling_mistakes}

    \caption{Performance vs. profile assignment method. \textbf{---Ablation 8}}
    \vspace{0.22cm}
    \resizebox{1\linewidth}{!}{%
    \begin{tabular}{c|cccc} 
    \hline
    \multirow{2}{*}{\begin{tabular}[c]{@{}c@{}}Profile Assig-\\nment Method \end{tabular}} 
    & \multirow{2}{*}{\begin{tabular}[c]{@{}c@{}}CIFAR\\-10N\end{tabular}} & \multirow{2}{*}{\begin{tabular}[c]{@{}c@{}}CIFAR\\-10H\end{tabular}} & \multirow{2}{*}{\begin{tabular}[c]{@{}c@{}}F-MNI-\\ST-H\end{tabular}} & \multirow{2}{*}{\begin{tabular}[c]{@{}c@{}}Chaoy-\\ang\end{tabular}} \\ 
     &  &  &  &  \\ 
    \hline\hline
    Soft & 0.982 & 0.989 & 0.866       & 0.945     \\
    Hard & 0.989 & 0.993 & 0.878       & 0.988     \\
    \hline
\end{tabular}
    }
    \label{table:abl_soft_assignements}
\end{minipage}
\end{table*}


\subsection{Setup for Datasets with Real Annotators}
When training with CIFAR-10N, we present two experiments.
For the first experiment, the labels from 747 annotators form  $\tilde{\mathcal{D}}$.
Out of them, 159 were identified for having annotated at least 20 images per class, and they were split in half, taking 79 as training users and 80 as testing users. 
The training users' labels are used to build the $K$ profiles and train the OVA SVM classifier, where $K$ is automatically chosen based on the silhouette score in  eq. \ref{eq:silhouette}.
During testing, a testing user's noisy-label transition matrix is estimated using the test annotator's labels and consensus labels.
This matrix is used to simulate noisy annotations from that testing user.
Therefore, 80 noisy test sets are produced, with each representing the biases that user possesses.
The model for each profile $k$, denoted by $m_{\theta_{k}}(.)$, uses ViT-Base-16 \cite{vit} as the $f_{\psi_{k}}(.)$. 

For the second CIFAR-10N experiment, we use CIFAR-10H as the testing set, where the labels from testing users were used without any modification for user profiling. The same labels were used to estimate a noise transition matrix and simulate their own test set. For all 2571 users, their own test test was simulated that possess own biases.
The models trained for CIFAR-10N were used for this experiment. 

For the Fashion-MNIST-H experiment, the labels from all 885 annotators are taken to form the $\tilde{\mathcal{D}}$. 
Then, 366 out of 885 users are chosen since they have annotated at least 20 images per class and are split in half to have 183 users for training and 183 for testing. 
The training users' labels are used to build the $K$ profiles and train the OVA SVM classifier, where $K$ is automatically chosen based on the silhouette score  in  eq.\ref{eq:silhouette}.
During testing, the testing user's noisy-label transition matrix is estimated using the annotator’s labels and consensus labels.
This matrix is used to simulate test annotations from that testing user.
Therefore, 183 noisy testing sets are produced, with each representing the biases that each user possesses.
The model for each profile $k$, represented by $m_{\theta_{k}}(.)$  uses DenseNet-121 \cite{dense121} for $f_{\psi_{k}}(.)$.

Chaoyang has three annotators per image, which form the $\tilde{\mathcal{D}}$. 
Training users are used to make $K$ profiles, and train an OVA SVM, where $K$ is automatically chosen based on the silhouette score in  eq.\ref{eq:silhouette}. For each profile $k$, a model $m_{\theta_{k}}(.)$ is trained with a ViT-Large-16 as the backbone for $f_{\psi_{k}}(.)$. 
During testing, user's noisy-label transition matrix is estimated using the annotator’s labels and consensus labels. This matrix is used to simulate noisy test annotations from that user, resulting three noisy test sets.

Our experiment with CIFAR-10N and CIFAR-10H, with human labels for CIFAR-10's training and testing sets respectively, offer a more realistic setup, better reflecting real-world conditions. But, while our method preserves annotators' noisy label patterns, it's important to note that Fashion-MNIST-H and Chaoyang test sets are simulated and might not completely mimic real annotator inputs.


In our CIFAR experiments, we adopted the data augmentation policy introduced by \cite{cifar_AutoAug}. Also, for Fashion-MNIST, alongside random horizontal and vertical flips, we integrated auto augmentations as proposed by \cite{fmnist_randaug}. For the Chaoyang dataset, data augmentation was limited to random resized crops of dimensions $224\times224$. 
We rely on pre-trained models for $f_{\psi_{k}}$ because of their robustness to noisy labels~\cite{pretraining_nl} (e.g., ViT models, ResNet-18 and DenseNet-121 models are pre-trained on ImageNet-1K.) 
Adam optimizer was employed for training $f_{\psi_{k}}(.)$ with consensus $\bar{\mathcal{D}}$, where NAdam was used for training $m_{\theta_{k}}(.)$ on $\hat{\mathcal{D}}$, each utilizing their respective default learning rates.
Implementations were done in PyTorch and executed on single GeForce RTX 4090 GPU.

\subsection{Setup for Experiment in Text Domain}
This experiment evaluates \projname{} in the text domain using AgNews \cite{agnews} following a simulation setup similar to CIFAR-10.
AgNews is a text classification dataset with 120,000 training, 200 validation and 7,400 testing class-balanced news articles categorized into 4 classes.
We perform pairwise label flipping on two out of four classes, where $60\%$ of samples are flipped to the incorrect class while the other two classes remained clean.
We simulate three profiles of users, one that flips between classes business$\leftrightarrow$science/technology, another profile that flips world$\leftrightarrow$sports, and the third profile that flips sports$\leftrightarrow$business.
Five training and five testing users are simulated for each profile producing a total of 15 unique training and testing users.
The title and description were concatenated and truncated to a maximum of 64 tokens, Bert-Tokenizer was used to tokenize and Bert-Base-Uncased \cite{bert} model is used as base model $f_{\psi_{k}}(.)$ when training.
All 15 test users showed improved post-alteration accuracy over original and a high positive alteration proportion, indicating while further future experiments are needed, adapting to the text domain is a possibility.

\begin{figure*}[tb]
    \centering
    \includegraphics[width=\linewidth]{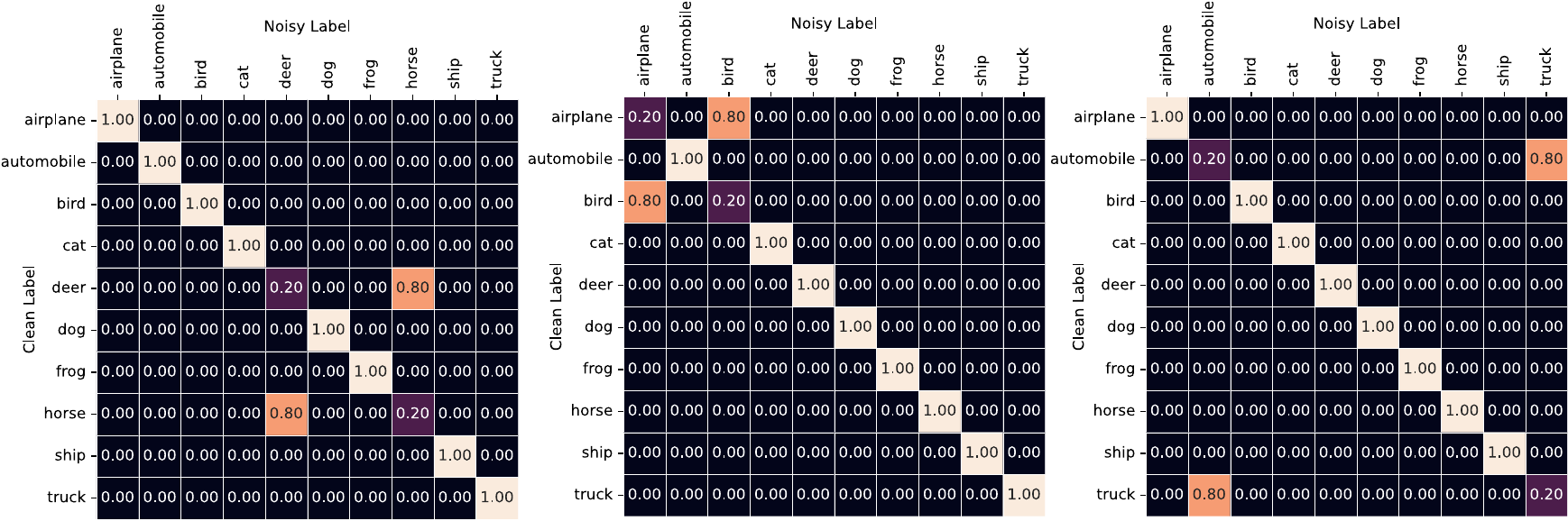}
    \caption{Estimated noise matrices for identified annotator profiles from CIFAR-10 simulation experiment.}
\label{fig:n_patterns_c10}
\end{figure*}

\section{\break Additional Ablation Studies}
\label{supp:additional_abl}
This section provides the results for four additional ablation studies: 
(1) post alteration accuracy vs. having a different $K$ than the optimal from the silhouette score (Ablation 5 in Table \ref{table:abl_K}),
(2) post alteration accuracy vs. $\lambda$ in loss function eq.\ref{eq:model_optimisation} (Ablation 6 in Table \ref{table:abl_lambda}),
(3) post alteration accuracy vs. SVM profiling errors (Ablation 7 in Table \ref{table:svm_profiling_mistakes}), and
(4) post alteration accuracy vs. profile assignment method being soft or hard (Ablation 8 in Table \ref{table:abl_soft_assignements})
; that we discussed in Sec. \ref{sec:ablation} in the main paper.

\section{\break Visualizing Noise profiles}
\label{supp:vis_noise_profiles}
Figure \ref{fig:n_patterns_c10} shows the estimated noise matrices of identified annotator profiles from the CIFAR-10 simulation with $K=3$, selected via the silhouette score.
Notably, these matrices closely resembles the simulated noise patterns used to create 15 users—flipping between airplane$\leftrightarrow$bird, horse$\leftrightarrow$deer, and truck$\leftrightarrow$automobile. 
This validates the ability of the clustering to capture distinct noise patterns and confirms the effectiveness of the silhouette score in selecting an optimal $K$ for accurate user profiling.

\section{\break Extended Related Work}
\label{supp:related_work_ex}
While restating that the goal of \projname{} is improving human-AI joint decision-making with unseen users, we review related yet distinct research areas due to the interdisciplinary nature of the problem.

Learning from Noisy-labels (LNL) aims to design algorithms that are robust to the presence of noisy training labels.
Recent advancements include
DivideMix ~\cite{nl_dividemix} with its semi-supervised approach,
ELR ~\cite{nl_elr} exploring early learning phenomena with a regularised loss, 
CausalNL ~\cite{yao2021causalnl} proposing a generative model for instance-dependent label-noise learning, 
C2D ~\cite{nl_c2d} tackling the warm-up obstacle, UNICON ~\cite{nl_unicon} with a unified supervised and unsupervised learning, and
\cite{graphical_modeling} with graphical modeling for noise rate estimation to handle noisy labels effectively.
In contrast, \projname{} does not aim to infer clean labels, but to improve human-AI joint decision-making by adapting to annotator-specific biases—shifting the goal from label denoising to cooperation.

Multi-rater Learning  (MRL) leverages noisy labels from multiple annotators per sample, often to estimate consensus or to mitigate the identifiability problem under certain conditions~\cite{liu2023identifiability}. 
Key developments include,
\cite{fast_dawidskene}, with expectation maximization algorithm to estimate ground truth,
MRNet~\cite{ji2021learning}, which addresses multi-rater disagreement, 
\cite{crowdlab}, estimating consensus being model-agnostic in design,
\cite{admoe} adopting a Mixture of Experts architecture for learning from multiple noisy sources,
and 
\cite{mrlcoupled} addressing the sparse crowd annotations.
We employ \cite{crowdlab} to estimate consensus in \projname{} for its model-agnostic nature and good performance, although it is an open choice in \projname{} framework.
Although consensus estimation is a preliminary step, \projname{}'s core contribution lies in complementing annotator-specific behavior—an objective different from MRL.

Learning with crowds (LWC) aims to train models with sparse labels provided by multiple annotators.
\cite{crowd_union} and \cite{crowd_colearning} aim to aggregate diverse annotator signals while addressing label reliability. 
Sel-CL \cite{crowd_selective} tackles noisy supervision by leveraging contrastive learning to better separate clean and corrupted labels.
AIDTM \cite{crowd_transferring} learns annotator-specific noise patterns, while CCC \cite{crowd_coupled} handles settings with limited labels per annotator through parameter-efficient modeling.
However, unlike LWC, that aims to denoise annotations, \projname{} preserves individual annotator signals to train profile-specific cooperative AI models that can adapt to and complement unseen users.

Despite improvements from LNL, MRL and LWC, an accuracy gap persists compared to training with clean labels. This has led to our human-AI joint decision-making paradigm, which incorporates inputs from both humans and AI to make decisions.
Unlike objectives of recommendation systems, LWC, LNL or MRL—that focus on preference modeling, inferring latent ground truth, being robust to label noise, or consensus estimation—our goal is to improve human-AI joint decision-making by complementing unseen users—the fundamental distinction of \projname{}.

\bibliographystyle{IEEEtran}
\bibliography{references}

@article{yao2021causalnl,
  title={Instance-dependent label-noise learning under a structural causal model},
  author={Yao, Yu and Liu, Tongliang and Gong, Mingming and Han, Bo and Niu, Gang and Zhang, Kun},
  journal={Advances in Neural Information Processing Systems},
  volume={34},
  pages={4409--4420},
  year={2021}
}

@inproceedings{dehariya2010clustering,
  title={Clustering of image data set using k-means and fuzzy k-means algorithms},
  author={Dehariya, Vinod Kumar and Shrivastava, Shailendra Kumar and Jain, RC},
  booktitle={2010 International conference on computational intelligence and communication networks},
  pages={386--391},
  year={2010},
  organization={IEEE}
}

@inproceedings{ji2021learning,
  title={Learning calibrated medical image segmentation via multi-rater agreement modeling},
  author={Ji, Wei and Yu, Shuang and Wu, Junde and Ma, Kai and Bian, Cheng and Bi, Qi and Li, Jingjing and Liu, Hanruo and Cheng, Li and Zheng, Yefeng},
  booktitle={Proceedings of the IEEE/CVF Conference on Computer Vision and Pattern Recognition},
  pages={12341--12351},
  year={2021}
}

@article{song2022learning,
  title={Learning from noisy labels with deep neural networks: A survey},
  author={Song, Hwanjun and Kim, Minseok and Park, Dongmin and Shin, Yooju and Lee, Jae-Gil},
  journal={IEEE Transactions on Neural Networks and Learning Systems},
  year={2022},
  publisher={IEEE}
}

@INPROCEEDINGS{Wang2021-Brilliant,
  title     = "“Brilliant {AI} Doctor” in Rural Clinics: Challenges in
               {AI}-Powered Clinical Decision Support System Deployment",
  author    = "Wang, Dakuo and Wang, Liuping and Zhang, Zhan and Wang, Ding and
               Zhu, Haiyi and Gao, Yvonne and Fan, Xiangmin and Tian, Feng",
  booktitle = "Proceedings of the 2021 CHI Conference on Human Factors in
               Computing Systems",
  publisher = "Association for Computing Machinery",
  address   = "New York, NY, USA",
  number    = "Article 697",
  pages     = "1--18",
  series    = "CHI '21",
  month     =  may,
  year      =  2021
}

@INPROCEEDINGS{Kocielnik2019-Imperfect,
  title     = "Will You Accept an Imperfect {AI}? Exploring Designs for
               Adjusting End-user Expectations of {AI} Systems",
  author    = "Kocielnik, Rafal and Amershi, Saleema and Bennett, Paul N",
  booktitle = "Proceedings of the 2019 CHI Conference on Human Factors in
               Computing Systems",
  publisher = "Association for Computing Machinery",
  address   = "New York, NY, USA",
  number    = "Paper 411",
  pages     = "1--14",
  series    = "CHI '19",
  month     =  may,
  year      =  2019
}

@ARTICLE{Strauch2018-IronyAutomation,
  author={Strauch, Barry},
  journal={IEEE Transactions on Human-Machine Systems}, 
  title={Ironies of Automation: Still Unresolved After All These Years}, 
  year={2018},
  volume={48},
  number={5},
  pages={419-433},
  doi={10.1109/THMS.2017.2732506}}

@article{raghu19_triage,
  title={The algorithmic automation problem: Prediction, triage, and human effort},
  author={Raghu, Maithra and Blumer, Katy and Corrado, Greg and Kleinberg, Jon and Obermeyer, Ziad and Mullainathan, Sendhil},
  journal={arXiv preprint arXiv:1903.12220},
  year={2019}
}

@inproceedings{liu2023identifiability,
  title={Identifiability of label noise transition matrix},
  author={Liu, Yang and Cheng, Hao and Zhang, Kun},
  booktitle={International Conference on Machine Learning},
  pages={21475--21496},
  year={2023},
  organization={PMLR}
}

@article{Vodrahalli2022_uncalibModels,
title={Uncalibrated models can improve human-ai collaboration},
author={Vodrahalli, Kailas and Gerstenberg, Tobias and Zou, James Y},
journal={Advances in Neural Information Processing Systems},
volume={35},
pages={4004--4016},
year={2022}
}

@article{Kerrigan2021-calibrateconfmatrix,
title={Combining human predictions with model probabilities via confusion matrices and calibration},
author={Kerrigan, Gavin and Smyth, Padhraic and Steyvers, Mark},
journal={Advances in Neural Information Processing Systems},
volume={34},
pages={4421--4434},
year={2021}
}

@InProceedings{whoshould_mozannar23,
  title = 	 {Who Should Predict? Exact Algorithms For Learning to Defer to Humans},
  author =       {Mozannar, Hussein and Lang, Hunter and Wei, Dennis and Sattigeri, Prasanna and Das, Subhro and Sontag, David},
  booktitle = 	 {Proceedings of The 26th International Conference on Artificial Intelligence and Statistics},
  pages = 	 {10520--10545},
  year = 	 {2023},
  editor = 	 {Ruiz, Francisco and Dy, Jennifer and van de Meent, Jan-Willem},
  volume = 	 {206},
  series = 	 {Proceedings of Machine Learning Research},
  month = 	 {25--27 Apr},
  publisher =    {PMLR},
  url = 	 {https://proceedings.mlr.press/v206/mozannar23a.html},
}

@INPROCEEDINGS{consistentest_Mozannar2020,
  title     = "Consistent Estimators for Learning to Defer to an Expert",
  booktitle = "Proceedings of the 37th International Conference on Machine
               Learning",
  author    = "Mozannar, Hussein and Sontag, David",
  editor    = "Iii, Hal Daum{\'e} and Singh, Aarti",
  publisher = "PMLR",
  volume    =  119,
  pages     = "7076--7087",
  series    = "Proceedings of Machine Learning Research",
  year      =  2020
}

@inproceedings{complement_wilder,
author = {Wilder, Bryan and Horvitz, Eric and Kamar, Ece},
title = {Learning to Complement Humans},
year = {2021},
isbn = {9780999241165},
booktitle = {Proceedings of the Twenty-Ninth International Joint Conference on Artificial Intelligence},
articleno = {212},
numpages = {8},
location = {Yokohama, Yokohama, Japan},
series = {IJCAI'20}
}

@ARTICLE{Bayesian_Steyvers22,
  title    = "Bayesian modeling of {human-AI} complementarity",
  author   = "Steyvers, Mark and Tejeda, Heliodoro and Kerrigan, Gavin and
              Smyth, Padhraic",
  journal  = "Proceedings of the National Academy of Sciences of the United States of America",
  volume   =  119,
  number   =  11,
  pages    = "e2111547119",
  month    =  mar,
  year     =  2022,
  language = "en"
}

@inproceedings{predictResponsibly_madras,
author = {Madras, David and Pitassi, Toniann and Zemel, Richard},
title = {Predict Responsibly: Improving Fairness and Accuracy by Learning to Defer},
year = {2018},
publisher = {Curran Associates Inc.},
address = {Red Hook, NY, USA},
booktitle = {Proceedings of the 32nd International Conference on Neural Information Processing Systems},
pages = {6150–6160},
numpages = {11},
location = {Montr\'{e}al, Canada},
series = {NIPS'18}
}

@article{crowdlab,
      title={CROWDLAB: Supervised learning to infer consensus labels and quality scores for data with multiple annotators}, 
      author={Hui Wen Goh and Ulyana Tkachenko and Jonas Mueller},
      year={2022},
      journal={NeurIPS 2022 Human in the Loop Learning Workshop},
}

@InProceedings{verma22OVA,
  title = 	 {Calibrated Learning to Defer with One-vs-All Classifiers},
  author =       {Verma, Rajeev and Nalisnick, Eric},
  booktitle = 	 {Proceedings of the 39th International Conference on Machine Learning},
  pages = 	 {22184--22202},
  year = 	 {2022},
  editor = 	 {Chaudhuri, Kamalika and Jegelka, Stefanie and Song, Le and Szepesvari, Csaba and Niu, Gang and Sabato, Sivan},
  volume = 	 {162},
  series = 	 {Proceedings of Machine Learning Research},
  month = 	 {17--23 Jul},
  publisher =    {PMLR},
  url = 	 {https://proceedings.mlr.press/v162/verma22c.html},
}

@article{okati21DiffTriage,
  title={Differentiable learning under triage},
  author={Okati, Nastaran and De, Abir and Rodriguez, Manuel},
  journal={Advances in Neural Information Processing Systems},
  volume={34},
  pages={9140--9151},
  year={2021}
}

@InProceedings{lossTerm,
author = {Patrini, Giorgio and Rozza, Alessandro and Krishna Menon, Aditya and Nock, Richard and Qu, Lizhen},
title = {Making Deep Neural Networks Robust to Label Noise: A Loss Correction Approach},
booktitle = {Proceedings of the IEEE Conference on Computer Vision and Pattern Recognition (CVPR)},
month = {July},
year = {2017}
}

@TECHREPORT{c10,
    author = {Alex Krizhevsky},
    title = {Learning multiple layers of features from tiny images},
    institution = {University of Toronto, Toronto, Ontario},
    year = {2009}
}

@ARTICLE{chaoyang,
  author={Zhu, Chuang and Chen, Wenkai and Peng, Ting and Wang, Ying and Jin, Mulan},
  journal={IEEE Transactions on Medical Imaging}, 
  title={Hard Sample Aware Noise Robust Learning for Histopathology Image Classification}, 
  year={2022},
  volume={41},
  number={4},
  pages={881-894},
  doi={10.1109/TMI.2021.3125459}}

@inproceedings{c10n,
    title={Learning with Noisy Labels Revisited: A Study Using Real-World Human Annotations},
    author={Jiaheng Wei and Zhaowei Zhu and Hao Cheng and Tongliang Liu and Gang Niu and Yang Liu},
    booktitle={International Conference on Learning Representations},
    year={2022},
    url={https://openreview.net/forum?id=TBWA6PLJZQm}
}

@InProceedings{c10h,
    author = {Peterson, Joshua C. and Battleday, Ruairidh M. and Griffiths, Thomas L. and Russakovsky, Olga},
    title = {Human Uncertainty Makes Classification More Robust},
    booktitle = {Proceedings of the IEEE/CVF International Conference on Computer Vision (ICCV)},
    month = {October},
    year = {2019}
}

@inproceedings{fmnisth,
  author={Takashi Ishida and Ikko Yamane and Nontawat Charoenphakdee and Gang Niu and Masashi Sugiyama},
  title={Is the Performance of My Deep Network Too Good to Be True? A Direct Approach to Estimating the Bayes Error in Binary Classification},
  year={2023},
  cdate={1672531200000},
  url={https://openreview.net/forum?id=FZdJQgy05rz},
  booktitle={ICLR},
}

@article{fmnist,
  title={Fashion-mnist: a novel image dataset for benchmarking machine learning algorithms},
  author={Xiao, Han and Rasul, Kashif and Vollgraf, Roland},
  journal={arXiv preprint arXiv:1708.07747},
  year={2017}
}

@inproceedings{dense121,
  title={Densely connected convolutional networks},
  author={Huang, Gao and Liu, Zhuang and Van Der Maaten, Laurens and Weinberger, Kilian Q},
  booktitle={Proceedings of the IEEE conference on computer vision and pattern recognition},
  pages={4700--4708},
  year={2017}
}

@inproceedings{
vit,
title={An Image is Worth 16x16 Words: Transformers for Image Recognition at Scale},
author={Alexey Dosovitskiy and Lucas Beyer and Alexander Kolesnikov and Dirk Weissenborn and Xiaohua Zhai and Thomas Unterthiner and Mostafa Dehghani and Matthias Minderer and Georg Heigold and Sylvain Gelly and Jakob Uszkoreit and Neil Houlsby},
booktitle={International Conference on Learning Representations},
year={2021},
url={https://openreview.net/forum?id=YicbFdNTTy}
}

@inproceedings{resnets,
  title={Deep residual learning for image recognition},
  author={He, Kaiming and Zhang, Xiangyu and Ren, Shaoqing and Sun, Jian},
  booktitle={Proceedings of the IEEE conference on computer vision and pattern recognition},
  pages={770--778},
  year={2016}
}

@inproceedings{pretraining_nl,
  title={Beyond synthetic noise: Deep learning on controlled noisy labels},
  author={Jiang, Lu and Huang, Di and Liu, Mason and Yang, Weilong},
  booktitle={International conference on machine learning},
  pages={4804--4815},
  year={2020},
  organization={PMLR}
}

@inproceedings{cifar_AutoAug,
  title={Autoaugment: Learning augmentation strategies from data},
  author={Cubuk, Ekin D and Zoph, Barret and Mane, Dandelion and Vasudevan, Vijay and Le, Quoc V},
  booktitle={Proceedings of the IEEE/CVF conference on computer vision and pattern recognition},
  pages={113--123},
  year={2019}
}

@inproceedings{fmnist_randaug,
  title={Randaugment: Practical automated data augmentation with a reduced search space},
  author={Cubuk, Ekin D and Zoph, Barret and Shlens, Jonathon and Le, Quoc V},
  booktitle={Proceedings of the IEEE/CVF conference on computer vision and pattern recognition workshops},
  pages={702--703},
  year={2020}
}

@inproceedings{nl_dividemix,
title={DivideMix: Learning with Noisy Labels as Semi-supervised Learning},
author={Junnan Li and Richard Socher and Steven C.H. Hoi},
booktitle={International Conference on Learning Representations},
year={2020},
url={https://openreview.net/forum?id=HJgExaVtwr}
}

@article{nl_elr,
  title={Early-learning regularization prevents memorization of noisy labels},
  author={Liu, Sheng and Niles-Weed, Jonathan and Razavian, Narges and Fernandez-Granda, Carlos},
  journal={Advances in neural information processing systems},
  volume={33},
  pages={20331--20342},
  year={2020}
}

@INPROCEEDINGS{nl_c2d,
  title     = "Contrast to divide: Self-supervised pre-training for learning
               with noisy labels",
  booktitle = "Proceedings of the {IEEE/CVF} Winter Conference on Applications
               of Computer Vision",
  author    = "Zheltonozhskii, Evgenii and Baskin, Chaim and Mendelson, Avi and
               Bronstein, Alex M and Litany, Or",
  pages     = "1657--1667",
  year      =  2022
}

@INPROCEEDINGS{nl_unicon,
  title     = "Unicon: Combating label noise through uniform selection and
               contrastive learning",
  booktitle = "Proceedings of the {IEEE/CVF} Conference on Computer Vision and
               Pattern Recognition",
  author    = "Karim, Nazmul and Rizve, Mamshad Nayeem and Rahnavard, Nazanin
               and Mian, Ajmal and Shah, Mubarak",
  pages     = "9676--9686",
  year      =  2022
}

@article{fifar,
  title={FiFAR: A Fraud Detection Dataset for Learning to Defer},
  author={Alves, Jean V and Leit{\~a}o, Diogo and Jesus, S{\'e}rgio and Sampaio, Marco OP and Saleiro, Pedro and Figueiredo, M{\'a}rio AT and Bizarro, Pedro},
  journal={arXiv preprint arXiv:2312.13218},
  year={2023}
}

@inproceedings{agnews,
  title={Character-level Convolutional Networks for Text Classification},
  author={Xiang Zhang and Junbo Jake Zhao and Yann LeCun},
  booktitle={NIPS},
  year={2015}
}

@article{bert,
  author    = {Jacob Devlin and
               Ming{-}Wei Chang and
               Kenton Lee and
               Kristina Toutanova},
  title     = {{BERT:} Pre-training of Deep Bidirectional Transformers for Language
               Understanding},
  journal   = {CoRR},
  volume    = {abs/1810.04805},
  year      = {2018},
  url       = {http://arxiv.org/abs/1810.04805},
  archivePrefix = {arXiv},
  eprint    = {1810.04805},
  bibsource = {dblp computer science bibliography, https://dblp.org}
}

@misc{fast_dawidskene,
      title={Fast Dawid-Skene: A Fast Vote Aggregation Scheme for Sentiment Classification}, 
      author={Vaibhav B Sinha and Sukrut Rao and Vineeth N Balasubramanian},
      year={2018},
      eprint={1803.02781},
      archivePrefix={arXiv},
      primaryClass={stat.ML}
}

@inproceedings{fuzKmean_robust,
author = {Xu, Jinglin and Han, Junwei and Xiong, Kai and Nie, Feiping},
title = {Robust and sparse fuzzy K-means clustering},
year = {2016},
isbn = {9781577357704},
publisher = {AAAI Press},
booktitle = {Proceedings of the Twenty-Fifth International Joint Conference on Artificial Intelligence},
pages = {2224–2230},
numpages = {7},
location = {New York, New York, USA},
series = {IJCAI'16}
}

@inproceedings{mrlcoupled,
  title={Coupled confusion correction: Learning from crowds with sparse annotations},
  author={Zhang, Hansong and Li, Shikun and Zeng, Dan and Yan, Chenggang and Ge, Shiming},
  booktitle={Proceedings of the AAAI Conference on Artificial Intelligence},
  volume={38},
  pages={16732--16740},
  year={2024}
}

@article{imgnet,
Author = {Olga Russakovsky and Jia Deng and Hao Su and Jonathan Krause and Sanjeev Satheesh and Sean Ma and Zhiheng Huang and Andrej Karpathy and Aditya Khosla and Michael Bernstein and Alexander C. Berg and Li Fei-Fei},
Title = {{ImageNet Large Scale Visual Recognition Challenge}},
Year = {2015},
journal   = {International Journal of Computer Vision (IJCV)},
doi = {10.1007/s11263-015-0816-y},
volume={115},
number={3},
pages={211-252}
}

@inproceedings{l2d_2pop,
  title={Learning to defer to a population: A meta-learning approach},
  author={Tailor, Dharmesh and Patra, Aditya and Verma, Rajeev and Manggala, Putra and Nalisnick, Eric},
  booktitle={International Conference on Artificial Intelligence and Statistics},
  pages={3475--3483},
  year={2024},
  organization={PMLR}
}

@article{ea_l2d,
  title={Expert-Agnostic Learning to Defer},
  author={Strong, Joshua and Saha, Pramit and Ibrahim, Yasin and Ouyang, Cheng and Noble, Alison},
  journal={arXiv preprint arXiv:2502.10533},
  year={2025}
}

@inproceedings{prob_l2d,
title={Probabilistic Learning to Defer: Handling Missing Expert Annotations and Controlling Workload Distribution},
author={Cuong C. Nguyen and Thanh-Toan Do and Gustavo Carneiro},
booktitle={The Thirteenth International Conference on Learning Representations},
year={2025},
url={https://openreview.net/forum?id=zl0HLZOJC9}
}

@article{l2cmh,
  title={Learning to Complement with Multiple Humans},
  author={Zhang, Zheng and Nguyen, Cuong and Wells, Kevin and Do, Thanh-Toan and Carneiro, Gustavo},
  journal={arXiv preprint arXiv:2311.13172},
  year={2023}
}

@inproceedings{l2cdmu,
  title={Learning to Complement and to Defer to Multiple Users},
  author={Zhang, Zheng and Ai, Wenjie and Wells, Kevin and Rosewarne, David and Do, Thanh-Toan and Carneiro, Gustavo},
  booktitle={European Conference on Computer Vision},
  pages={144--162},
  year={2024},
  organization={Springer}
}

@inproceedings{graphical_modeling,
  title={Instance-Dependent Noisy-Label Learning with Graphical Model Based Noise-Rate Estimation},
  author={Garg, Arpit and Nguyen, Cuong and Felix, Rafael and Do, Thanh-Toan and Carneiro, Gustavo},
  booktitle={European Conference on Computer Vision},
  pages={372--389},
  year={2024},
  organization={Springer}
}

@inproceedings{admoe,
  title={Admoe: Anomaly detection with mixture-of-experts from noisy labels},
  author={Zhao, Yue and Zheng, Guoqing and Mukherjee, Subhabrata and McCann, Robert and Awadallah, Ahmed},
  booktitle={Proceedings of the AAAI Conference on Artificial Intelligence},
  volume={37},
  pages={4937--4945},
  year={2023}
}

@ARTICLE{crowd_union,
  author={Wei, Hongxin and Xie, Renchunzi and Feng, Lei and Han, Bo and An, Bo},
  journal={IEEE Transactions on Neural Networks and Learning Systems}, 
  title={Deep Learning From Multiple Noisy Annotators as A Union}, 
  year={2023},
  volume={34},
  number={12},
  pages={10552-10562},
  doi={10.1109/TNNLS.2022.3168696}
}

@article{crowd_colearning,
  title={Trustable co-label learning from multiple noisy annotators},
  author={Li, Shikun and Liu, Tongliang and Tan, Jiyong and Zeng, Dan and Ge, Shiming},
  journal={IEEE Transactions on Multimedia},
  volume={25},
  pages={1045--1057},
  year={2021},
  publisher={IEEE}
}

@inproceedings{crowd_selective,
  title={Selective-supervised contrastive learning with noisy labels},
  author={Li, Shikun and Xia, Xiaobo and Ge, Shiming and Liu, Tongliang},
  booktitle={Proceedings of the IEEE/CVF conference on computer vision and pattern recognition},
  pages={316--325},
  year={2022}
}

@article{crowd_transferring,
  title={Transferring annotator-and instance-dependent transition matrix for learning from crowds},
  author={Li, Shikun and Xia, Xiaobo and Deng, Jiankang and Ge, Shiming and Liu, Tongliang},
  journal={IEEE Transactions on Pattern Analysis and Machine Intelligence},
  volume={46},
  number={11},
  pages={7377--7391},
  year={2024},
  publisher={IEEE}
}

@inproceedings{crowd_coupled,
  title={Coupled confusion correction: Learning from crowds with sparse annotations},
  author={Zhang, Hansong and Li, Shikun and Zeng, Dan and Yan, Chenggang and Ge, Shiming},
  booktitle={Proceedings of the AAAI conference on artificial intelligence},
  volume={38},
  pages={16732--16740},
  year={2024}
}

\newpage

\EOD

\end{document}